\documentclass{article}
\usepackage{ijcai26}

\usepackage{times}
\usepackage{soul}
\usepackage{url}
\usepackage[hidelinks]{hyperref}
\usepackage[utf8]{inputenc}
\usepackage[small]{caption}
\usepackage{graphicx}
\usepackage{amsmath}
\usepackage{booktabs}
\usepackage{amssymb}
\usepackage[switch]{lineno}

\usepackage{forest}
\usetikzlibrary{arrows.meta,shapes,positioning,trees}
\tikzset{
    root/.style={fill=orange!20, font=\footnotesize},
    main/.style={font=\footnotesize, text width=3cm, align=center,},
    A/.style={fill=yellow!10},
    B/.style={fill=red!10},
    C/.style={fill=blue!10},
    edge from parent/.style={draw=gray, edge from parent fork right}
}
\forestset{
  myedge/.style={edge path={\noexpand\path[\forestoption{edge},-, >={latex}] 
         (!u.south) -- +(5pt,0pt) |- (.child anchor)
         \forestoption{edge label};}
  }
}

\title{Machine Learning Methods for Studying Latent Neural Activity Dynamics}

\author{
    Shufeng Kong$^{1,2}$, Fumei Deng$^{1}$\thanks{These authors contributed equally.}, Xinyi Dong$^{1}$\footnotemark[1],  Caihua Liu\textsuperscript{5,4,2}\thanks{Correspondence to Caihua Liu: cl2869@cornell.edu}, Weiwei Chen$^3$, Yingheng Wang$^2$, Daniel Cao$^2$, Azahara Oliva$^3$, Antonio Fernández-Ruiz$^3$ and Carla Gomes$^2$
    \affiliations
    $^1$School of Software Engineering, Sun Yat-sen University, Zhuhai, China \\
    $^2$Department of Computer Science, Cornell University Ithaca, NY, USA \\
    $^3$Department of Neurobiology and Behavior, Cornell University, Ithaca, NY, USA \\
    $^4$Department of Ecology and Evolutionary Biology, Cornell University, Ithaca, NY, USA \\
    $^5$School of Computer Science and Artificial Intelligence, Foshan University, Foshan, China
    \emails
    \{sk2299, cl2869, wc677, yw2349, dyc33, aog35, afr77\}@cornell.edu, gomes@cs.cornell.edu 
}

\begin{document}

\maketitle

\begin{abstract}
Recent developments in brain recording are driving a demand for machine learning tools capable of decoding the latent structure of large populations of neurons.  In this paper, we provide a comprehensive survey\footnote{Project Page: \url{https://github.com/sk2299/neuro-ai-survey}} that outlines the trajectory of Latent Variable Models (LVMs) from early state-space models to more recent deep generative models. We organize the literature into three closely related domains: (1) Single-Region Latent Dynamics, which includes models such as linear dynamical systems to more complex dynamics represented by Recurrent Neural Networks (RNNs) and Neural Ordinary Differential Equations (ODEs); (2) Multi-Region Communication, which employs probabilistic as well as subspace methods to study how information is transferred across different brain areas considering synaptic propagation delays and network connectivity; and (3) Behavior-Aligned Modeling, which seeks to disentangle neural activity related to task performance from other internal states via supervised or contrastive learning. 
Finally, we conclude and discuss benchmarks, evaluation criteria, and open challenges, such as the ability to identify causal links or directionality of communication, to facilitate future research for bridging interpretable brain dynamics with reliable neural decoding.  
\end{abstract}

\section{Introduction}

Recording the activity of hundreds of neurons across brain areas in behaving animals is becoming commonplace in neuro science study thanks to the development of large-scale neural recording tools \cite{Urai2022}. This wealth of data has prompted a revolution on the way neural activity is analyzed, to a shift in focus from single neuron responses to understanding the dynamics of large populations. An important conceptual drive in the development of novel analytical and modeling approaches is the idea that complex, high-dimensional neural dynamics can be largely explained by a much reduced set of factors, or `latent variables' \cite{CEBRA}.
Therefore, the main goal for Artificial Intelligence (AI) in this domain is no longer just to analyze data, but to reverse-engineer the latent dynamical rules that govern how biological networks process information. To that end, there is a recent but growing appeals in the computational neuroscience community for Latent Variable Models (LVMs). Inspired by the neural manifold assumption \cite{gao2017theory}, which suggests that the activity of $N$ neurons is driven by a latent state $\mathbf{x}_t \in \mathbb{R}^K$ evolving on a low-dimensional manifold, a recent trend is to develop LVMs that satisfy dynamical, distributed, and behavioral criteria \cite{perich2025neural}.

In light of the rapid developments and emerging challenges of AI in neuroscience, we present a comprehensive survey of this field to help the community keep track of its progress. Specifically, we introduce a taxonomy that categorizes existing works into three phases: 
(1) \textbf{Single-Region Latent Dynamics} move from static smoothing to dynamical modeling. The fundamental idea is to leverage Recurrent Neural Networks (RNNs) \cite{LFADS} and Neural Ordinary Differential Equations (ODEs) \cite{PLNDE} to infer dynamical features such as attractors; 
(2) \textbf{Multi-Region Communication} builds on the success of single-region models to capture distributed interactions. To move a step forward, sophisticated subspace identification methods \cite{DLAG,CCA,MR-LFADS} have been used to show how information is routed between brain areas; 
(3) \textbf{Behavior-Aligned Modeling} integrates specialized learning frameworks to enhance behavioral decoding. Specifically, it applies contrastive learning \cite{CEBRA} and supervised disentanglement \cite{DPAD,PSID} to isolate latent factors that directly correlate with actions. 

Lastly, the field currently stands at the precipice of a fourth paradigm: Neural Foundation Models. Inspired by the success of Large Language Models (LLMs), recent works \cite{POYO,NEDS} treat neural spikes as tokens, aiming to learn a universal neural grammar that generalizes across individuals. 
\textcolor{black}{While existing reviews have focused on specific architectures, biological questions, or foundation models, we provide a unified perspective on latent variable methods for analyzing neural population dynamics, cross-regional interactions, and neuro-behavioral alignment within a common generative framework. We organize the literature into three complementary pillars: single-region latent dynamics, multi-region communication, and behavior-aligned modeling, offering a roadmap for ML researchers to engage with one of the most complex dynamical systems in nature.}

\begin{figure*}[htbp]
\centering
\resizebox{0.99\textwidth}{!}{%
\begin{forest} 
for tree={
    grow=east,
    minimum height=7mm,
    align=center,
    growth parent anchor=east,
    rectangle, draw, rounded corners=4pt,
    parent anchor=east,
    child anchor=west,
    edge path={\noexpand\path[\forestoption{edge},-, >={latex}] 
          (!u.parent anchor) -- +(5pt,0pt) |- (.child anchor)
          \forestoption{edge label};}
}
[Latent Neural Dynamics \& Communication, root, rotate=90,
    [Behavior-Aligned \\ Latent Modeling, main, C, myedge,
        [Generative \& Foundation Models, C 
            [Neuro-foundation Models, C[{POYO \cite{POYO},POSSM \cite{POSSM},NEDS \cite{NEDS}},C]]
            [Energy-Based \& Autoregressive Models, C[{EAG \cite{EAG}},C]]
            [Deep Generative Models, C[{LDNS \cite{LDNS}, BeNeDiff \cite{BeNeDiff}, GNOCCHI \cite{GNOCCHI}},C]]
        ]
        [Representation Alignment \& Geometry, C
            [Cross-Scenario Consistency Alignment, C[{SABLE \cite{SABLE}, NLA \cite{NLA}},C]]
            [Contrastive Learning \& Geometric Structure, C[{CEBRA \cite{CEBRA}, GLUE \cite{GLUE}},C]]
        ]
        [Disentangled Dynamic Modeling, C
            [Non-linear \& RNN-based Disentanglement, C[{DPAD \cite{DPAD}, BRAID \cite{BRAID}, LINT \cite{LINT}, SRNN \cite{SRNN}},C]]
            [Linear Subspace Disentanglement, C[{PSID \cite{PSID}, IPSID \cite{IPSID}},C]]
        ]
    ]
    [Multi-Region \\ Communication in \\ Latent Space, main, B, myedge,
        [Subspace Identification \& Alignment, B
            [Multi-view Alignment, B[{CREIMBO \cite{CREIMBO}},B]]
            [Communication Subspace Analysis, B[{CCA \cite{CCA}, RRR-based \cite{RRR-based}, Structured SNN + RRR \cite{SNN+RRR}},B]]
        ]
        [Deep Representation Learning Models, B
            [Network \& Geometric Dynamics, B[{mRNNs \cite{mRNN}, 
            FORCE \& full-FORCE \cite{fullFORCE}, SPLICE \cite{SPLICE}
            MARBLE \cite{MARBLE}},B]]
            [Sequential \& Transformer VAEs, B[{MR-LFADS \cite{MR-LFADS}, GLM-Transformer \cite{GLM-Transformer}, MtM \cite{MtM}},B]]
        ]
        [Probabilistic Generative Models, B
            [Switching \& Nonlinear SSMs, B[{MR-SDS \cite{MR-SDS}, mp-srSLDS \cite{mp-srSLDS}, MRDS-IR \cite{MRDS-IR}},B]]
            [Linear \& Gaussian Process Models, B[{DLAG \cite{DLAG}, mDLAG \cite{mDLAG}, MRM-GP \cite{MRM-GP}, ADM \cite{ADM}},B]]
        ]
    ]
    [Single-Region \\ Latent Dynamics, main, A, myedge,
        [Continuous \& Switching Dynamics, A 
            [Switching Linear Dynamical Systems, A [{rSLDS \cite{rSLDS}, gpSLDS \cite{gpSLDS}, DynCL \cite{laiz2024self}},A]]
            [Continuous-Time Neural ODEs \& SDEs, A [{PLNDE \cite{PLNDE}, FINDR \cite{kim2023flow}, ODIN \cite{ODIN}, LangevinFlow \cite{LangevinFlow}},A]]
        ]
        [Deep Sequential Generative Models, A
            [Transformer-based Models, A [{NDT \cite{NDT}},A]]
            [RNN-Based Variational Autoencoders, A [{LFADS \cite{LFADS}, AutoLFADS \cite{AutoLFADS}, DASC \cite{DASC}, Stochastic Low-Rank RNN \cite{SLRNN}},A]]
        ]
        [Probabilistic \& Linear State-Space Models, A
            [Linear Dynamical Systems, A[{PLDS \cite{PLDS}, CTDS \cite{CTDS}}, A]]
            [Gaussian Process-based, A[{GPFA \cite{GPFA}, vLGP \cite{vLGP}, SNP-GPFA \cite{SNP-GPFA}, P-GPLVM \cite{wu2017gaussian}},A]]
        ]
    ]
]
\end{forest}
}
\caption{A taxonomy of latent neural modeling. We categorize methods into: (1) Single-Region Dynamics (e.g., GPFA, RNNs), modeling intrinsic temporal evolution; (2) Multi-Region Communication (e.g., CCA, DLAG), capturing distributed information routing; and (3) Behavior-Aligned Modeling (e.g., CEBRA, PSID), using disentanglement for functional identification. The framework culminates in Neuro-Foundation Models (e.g., POYO), leveraging large-scale pre-training for neurobehavior decoding.}
\label{fig:taxonomy}
\end{figure*}
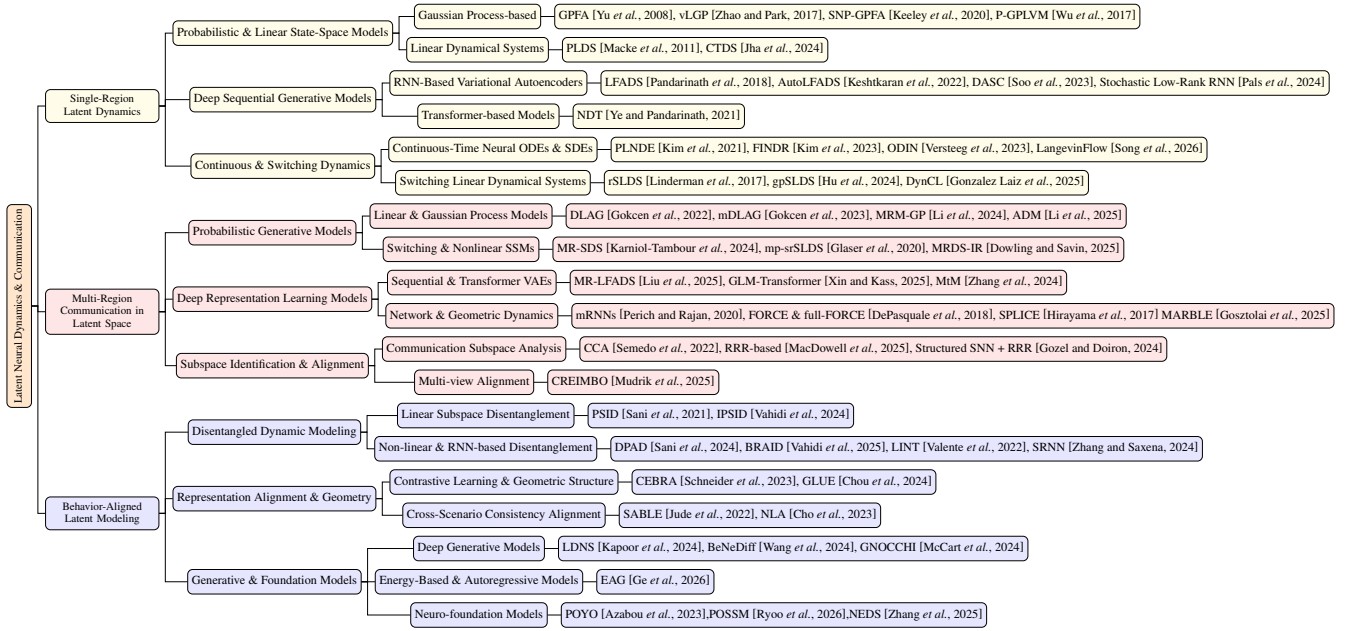

\section{Taxonomy and Preliminaries}

\subsection{Preliminaries and Problem Formulation}
To compare different latent variable approaches, we characterize neural population analysis as a generative process, which yields the general framework defined below and lets us classify methods by how they infer the latent neural states.

We generalize over specific architectural choices by regarding a neural recording as a sequence $\mathbf{Y}$ of $T$ observations. We can now formulate the generative process in terms of the governing equations as follows:
\begin{align}
    \mathbf{x}_{t} &= f(\mathbf{x}_{t-1}, \mathbf{u}_t, \epsilon_t) \label{eq:dynamics}, \\
    \mathbf{y}_{t} &\sim \mathcal{P}(g(\mathbf{x}_t, \mathbf{s}_t)) \label{eq:observation},
\end{align}
where $\mathbf{u}_t$ represents external inputs, and $\epsilon_t$ denotes process noise. The observation is drawn from a probability distribution $\mathcal{P}$, parameterized by the mapping function $g(\cdot)$. The term $\mathbf{s}_t$ accounts for optional history-dependent covariates. Existing work on latent variable models uses many different types of objective functions; some allow for only inference (estimating the posterior probability distribution), while others combine them using learning objectives (maximizing the likelihood).

\subsection{Taxonomy of Latent Variable Models}
We introduce our taxonomy and classification for the surveyed papers (see Figure~\ref{fig:taxonomy}) from the perspective of a model designer, focusing mainly on the interplay between the objective function and the neural dynamics.

\subsubsection{Single-Region Latent Dynamics}
The most basic level focuses on individual brain regions alone, characterizing the intrinsic rules and latent factors that govern intra-area dynamics. Although the absence of inter-regional coupling makes this seem simpler, the surveyed papers reveal subtle complexity that varies with the function $f$: Neural ODEs \cite{PLNDE} and RNNs \cite{LFADS} both model single-region dynamics, yet identify different computational primitives such as attractors or limit cycles.

\subsubsection{Multi-Region Communication}
Cognitive functions rely on distributed populations interacting across regions, so a primary goal is to understand how regions influence one another and route information---challenging because of the bidirectional loops pervading most areas. To separate local computation from information routing, we analyze the latent manifold extracted by $g$: shared latent components are classified as multi-region communication, while region-private components are local computation.

\subsubsection{Behavior-Aligned Modeling}
This category operates within supervised or behavior-guided paradigms \cite{DPAD,CEBRA}, where the objective function acts as a functional constraint: the validity of a representation is judged by its alignment with behavior (e.g., ``the latent space must disentangle motor output'' or ``the manifold must be invariant across subjects''). Such constraints effectively assign zero utility to any state $\mathbf{x}$ outside the set of behaviorally aligned states. We also examine large-scale pre-training for neurobehavioral modeling (e.g. Neuro-Foundation Models \cite{POYO,NEDS}) which target cross-subject generalization, zero-shot decoding, and multimodal integration.

\section{Single-Region Latent Dynamics}

\subsection{Probabilistic and Linear State-Space Models}
\subsubsection{Gaussian Process-based Models}
To handle the ill-posed nature of neural decoding, various state-space models have been developed. One of the first \cite{GPFA} defines the latent state evolution not through explicit dynamics, but via a kernel function that enforces temporal smoothness. Specifically, \textit{Gaussian-Process Factor Analysis} (GPFA) extracts neural trajectories by assuming the observed activity $\mathbf{y}_{:,t}$ is a linear projection of a latent state $\mathbf{x}_{:,t}$ driven by a Gaussian Process (GP) prior. The likelihood is given by:
\begin{equation}
\mathbf{y}_{:,t} \mid \mathbf{x}_{:,t} \sim \mathcal{N}\!\left( \mathbf{C} \mathbf{x}_{:,t} + \mathbf{d},\, \mathbf{R} \right).
\end{equation}
Much of the following work adds new inference strategies for non-Gaussian noise, since the standard GPFA assumption is ill-suited for discrete spike trains. The \textit{Variational Latent Gaussian Process} (vLGP) \cite{vLGP} addresses this with a variational framework using a Poisson observation model, while \textit{SNP-GPFA} \cite{SNP-GPFA} disentangles stimulus-locked signal from internal noise via Fourier-domain black-box variational inference. Finally, to overcome linear mappings, \textit{P-GPLVM} \cite{wu2017gaussian} employs GPs to govern both the temporal latent variable and non-linear tuning curves, discovering low-dimensional structure in complex manifolds such as hippocampal place-cell codes.

\subsubsection{Linear Dynamical Systems}
Unlike GPFA, \emph{Linear Dynamical Systems} (LDS) explicitly model the generative process via a state transition matrix $\mathbf{A}$:
\begin{equation}
\mathbf{x}_{t+1} = \mathbf{A} \mathbf{x}_t + \mathbf{B} \mathbf{u}_t + \mathbf{w}_t,
\end{equation}
where $\mathbf{u}_t$ represents external inputs. Then the Poisson Linear Dynamical System (PLDS) \cite{PLDS} combines this linear backbone with a Poisson emission model:
\begin{equation}
\mathbb{E}\!\left[ y^i_{t} \mid \mathbf{x}_{t} \right] = \exp\!\left( \left[ \mathbf{C} \mathbf{x}_{t} + \mathbf{d} \right]_i \right).
\end{equation}
While PLDS captures temporal evolution better than GLMs, standard LDS ignores biological constraints. \textit{CTDS} \cite{CTDS} enforces Dale's Law, which a neuron is purely excitatory or inhibitory, by restricting the connectivity columns to be non-negative, distinguishing excitatory from inhibitory populations and adding biophysical interpretability.

\subsection{Deep Sequential Generative Models}


\subsubsection{RNN-Based Variational Autoencoders}
This class shows strong recurrence between time steps. \emph{Latent Factor Analysis via Dynamical Systems} (LFADS) \cite{LFADS} is a key example: a Sequential Variational Autoencoder (SVAE) that infers the initial condition $g_0$ of a generator RNN. Then \textit{AutoLFADS} \cite{AutoLFADS} eliminates the need for expert hyperparameter tuning by integrating Population-Based Training (PBT) and Coordinated Dropout, which enables the model to self-optimize its architecture. \textit{DASC} \cite{DASC} further solves the gradient instability issues inherent in training leaky RNNs (which mimic membrane time constants) by introducing dynamically aligned skip connections. Finally, the \emph{Stochastic Low-Rank RNN} (SLRNN) \cite{SLRNN} injects interpretability into the black box of RNNs by constraining the recurrent matrix to a low-rank structure, which allows for analytical derivation of fixed points and flow fields.

\subsubsection{Transformer-Based Models}
The \emph{Neural Data Transformer} (NDT) \cite{NDT} adapts the Transformer to neuroscience to capture more complex temporal interdependencies. Instead of recurrence, NDT uses self-attention between temporal queries $Q$ and keys $K$ across the entire recording session. NDT utilizes a masked modeling objective (inspired by BERT), in which spikes are randomly masked and the model is trained to reconstruct them. This approach achieves faster inference than LFADS, demonstrating that explicit recurrence is not strictly necessary for recovering autonomous neural dynamics.

\subsection{Continuous \& Switching Dynamics}

\subsubsection{Continuous-Time Neural ODEs \& SDEs}

Continuous-time models prevent discretization errors. \textit{PLNDE} \cite{PLNDE} and \textit{ODIN} \cite{ODIN} utilize Neural Ordinary Differential Equations (ODEs) to capture deterministic evolution. Formally, these models define the derivative as:
\begin{equation}
  \dot{\mathbf{z}} = f_{\psi}(\mathbf{z}, t) + \sum_{u} \delta_{t_u}(\mathbf{g}^u_\phi + \eta_u).
\end{equation}

\textcolor{black}{
    To incorporate physical priors into continuous dynamics, LangevinFlow \cite{LangevinFlow} further models the evolution of neural latents as an underdamped Langevin equation. Within a sequential VAE framework, this model unifies intrinsic deterministic dynamics with unobserved stochastic influences:
\begin{equation}
    m\frac{\partial \mathbf{v}}{\partial t} = -\nabla_{\mathbf{z}} U(\mathbf{z}) - m\gamma \mathbf{v} + \sqrt{2 m \gamma k_B \tau} \boldsymbol{\eta}(t)
\end{equation}
To capture the oscillatory and flow-like characteristics inherent in neural activity, the potential function is innovatively parameterized as a locally coupled network of oscillators ($U(\mathbf{z}) = \mathbf{z}^T \frac{W_z}{||W_z||_2} \mathbf{z}$). Combined with a global Transformer decoder, the model can effectively disentangle and extract complex neural spatiotemporal dynamics.}

However, neural dynamics often exhibit intrinsic variability. \textit{FINDR} \cite{kim2023flow} extends this paradigm to \emph{Neural Stochastic Differential Equations} (SDEs). By parameterizing the drift with a Gated Neural Network and incorporating a diffusion term ($d\mathbf{z} = \mu(\mathbf{z}, \mathbf{u})dt + \sqrt{\Sigma}dW$), FINDR explicitly enables the visualization of latent {flow fields} and attractor structures, while simultaneously disentangling task-relevant dynamics from background noise.

\subsubsection{Switching Linear Dynamical Systems}

A \emph{Switching Linear Dynamical System} (SLDS) models complex dynamics by preserving linearity within individual modes; strictly speaking, the system's evolution remains linear as long as the underlying regime is maintained. An illustrative example is \textit{Recurrent Switching Linear Dynamical System} (rSLDS), which approximates a global nonlinear landscape by dynamically switching between distinct linear modes. Formally, this class of methods considers functions such that:
\begin{equation}
  \kappa_{ssl}(\mathbf{x}, \mathbf{x}') = \sum_{j=1}^J (\kappa_{lin}^{(j)}(\mathbf{x}, \mathbf{x}') \cdot \kappa_{part}^{(j)}(\mathbf{x}, \mathbf{x}')).
\end{equation}

In this framework, the switching kernel plays a pivotal role. As demonstrated in methods like \textit{gpSLDS} \cite{gpSLDS}, the kernel enables the model to iteratively refine the mode allocation, searching for optimal structural decompositions. 
Finally, recent approaches leverage contrastive learning to enforce identifiability. \textit{Dynamics Contrastive Learning} (DynCL) \cite{laiz2024self} introduces a framework that uncovers switching linear dynamics by optimizing an InfoNCE objective rather than a reconstruction loss. By differentiating through a bank of stable linear matrices, it identifies the underlying switching manifold without requiring the heavy computational overhead of generative RNNs.

\section{Multi-Region Communication in Latent Space}

\subsection{Probabilistic Generative Models}

These methods extend state-space models to capture non-stationary interactions and transmission delays between distinct neuronal populations.

\subsubsection*{Linear and Gaussian Process Models}
A fundamental challenge in multi-region modeling is decoupling shared variance from local variance. The unified observation process is typically expressed as:
\begin{equation}
\mathbf{y}_t = \mathbf{C} \begin{bmatrix} \mathbf{x}^{shared}_t \\ \mathbf{x}^{private}_t \end{bmatrix} + \mathbf{d} + \boldsymbol{\epsilon}_t,
\end{equation}
where latent states are partitioned into across-area variables $\mathbf{x}^{shared}$ and within-area variables $\mathbf{x}^{private}$.

\textit{DLAG} \cite{DLAG} and \textit{mDLAG} \cite{mDLAG} integrated a multi-output squared-exponential kernel to account for the physical transmission time of neural signals. By explicitly parameterizing the delay $\tau$ between regions $i$ and $j$ in the covariance function, these models effectively learn directed functional connectivity alongside signal propagation latencies.

However, simple time shifts often fail to capture frequency-specific changes. In response to this, \textit{MRM-GP} \cite{MRM-GP} employed complex-valued kernels to model phase delays. Furthermore, \textit{ADM} \cite{ADM} addresses non-stationary regimes by introducing a Markov Gaussian Process. Thanks to a time-varying transition matrix $\mathbf{A}_t$, the framework tracks how communication latencies evolve dynamically over the course of a single trial.

\subsubsection*{Switching and Nonlinear State-Space Models}
Communication between brain areas is not static; it is gated by internal cognitive states (e.g., attention). To capture this, researchers leverage \textit{Switching State-Space Models} (SSMs). The theoretical foundation for this approach is rSLDS \cite{rSLDS}. Although originally designed for general non-linear population dynamics, rSLDS introduced the critical mechanism of ``recurrence," where the discrete regime switch depends on the continuous latent history:
\begin{align}
    z_t &\mid z_{t-1}, \mathbf{x}_{t-1} 
        \sim \mathrm{Cat}\left(\pi(z_{t-1}, \mathbf{x}_{t-1})\right) \\
    \mathbf{x}_t &\mid \mathbf{x}_{t-1}, z_t
        \sim \mathcal{N}\!\left(f_{z_t}(\mathbf{x}_{t-1}),\, \mathbf{Q}_{z_t}\right).
\end{align}
In the context of distributed circuits, this generative structure is adapted to gate information flow rather than just local dynamics. For instance, \textit{MR-SDS} \cite{MR-SDS} extends the framework to multi-region settings, using the discrete state $z_t$ to toggle specific communication channels. Similarly, \textit{mp-srSLDS} \cite{mp-srSLDS} enhances the transition function $\pi(\cdot)$ with Neural Networks to model state-dependent stickiness, effectively capturing how neural populations sustain stable modes of communication over time.

\subsection{Deep Representation Learning Models}


\subsubsection{Sequential and Transformer VAEs}

To capture the nonlinear nature of neural communication, methods leverage deep representation learning. \textit{MR-LFADS} \cite{MR-LFADS} learns directed nonlinear influence via a Sender-Receiver architecture. Specifically, it infers a latent state from a source region that acts as a forcing input to a target region by assuming the instantaneous firing rate log-intensity $\log {\lambda}(t)$ is composed of additive functional components. The decomposition is given by:
\begin{equation}\log {\lambda}(t) = \mathcal{D}(\mathbf{x}_{local}) + \mathcal{C}(\mathbf{x}_{coupling}) + \mathcal{H}(\mathbf{y}_{history}),
\end{equation}
where $\mathcal{C}$ represents the population-coupling component.

Standard deep learning models often lack transparency regarding trial-specific factors. The \textit{GLM-Transformer} \cite{GLM-Transformer} addresses this by introducing a self-attention mechanism fused with \textit{Generalized Linear Model} (GLM) interpretability. On the other hand, communication dynamics often vary a lot across distinct recording environments. \textit{MtM} (Multi-task Masking) \cite{MtM} deals with these interdependencies by learning universal spatiotemporal representations via masked modeling, enabling generalization across different sessions and subjects.

\subsubsection{Network and Geometric Dynamics}

Methods in this category focus on the functional connectivity and intrinsic topology of the neural population. \textit{mRNNs} \cite{mRNN} fits large-scale time-series to a unified dynamical system, extracting a directed interaction matrix by assuming the evolution follows $\tau \dot{\mathbf{x}} = \mathcal{F}(\mathbf{x}, \mathbf{u}, \mathbf{\Theta})$. Complementing this inference approach, \textit{FORCE} and \textit{full-FORCE} \cite{fullFORCE} provide protocols for training the recurrent connectivity of Neural Networks to match target dynamics. While these are primarily tools for modeling the mechanisms of neural computation rather than inferring latent states, they offer a framework for understanding how complex dynamics emerge from recurrent architectures.

A distinct and emerging trend moves beyond temporal prediction to focus on the geometry of the neural manifold. \textit{SPLICE} \cite{SPLICE} pioneers this by introducing a hierarchical model with subspace pooling, learning representations that are invariant to local transformations (such as phase shifts). Building on this geometric perspective, \textit{MARBLE} \cite{MARBLE} addresses the sensitivity of models to local recording bases. It introduces an unsupervised geometric deep learning framework that decomposes dynamics into Local Flow Fields (LFFs), using contrastive learning to enforce invariance to local rotations and ensuring the learned dynamics are intrinsic to the underlying manifold.

\subsection{Subspace Identification and Alignment}
Methods for modeling interactions across distributed brain circuits must account for specific structural constraints to accurately capture inter-regional communication. Among these, the most critical requirements include achieving a low-rank dimensionality to isolate shared signals, maintaining designated orthogonality for factor disentanglement, and satisfying specific predictive targets between populations.

\subsubsection*{Communication Subspace Analysis}
\textit{Communication Subspace Analysis} (CSA) methods quantify the relationship between source activity $\mathbf{X}$ and target activity $\mathbf{Y}$ via a rank-constrained linear mapping:
\begin{equation}
\hat{\mathbf{Y}} \approx \mathbf{X} \mathbf{B}_{\mathrm{RRR}}, \quad \text{s.t. } \text{rank}(\mathbf{B}_{\mathrm{RRR}}) = m,
\end{equation}
where $m$ represents the dimensionality of the communication channel. \textit{Canonical Correlation Analysis} (CCA) \cite{CCA} identifies dimensions where source and target are maximally correlated. \textit{Reduced Rank Regression} (RRR) \cite{RRR-based} prioritizes predictive accuracy under rank constraints. \textit{Structured SNN + RRR} \cite{SNN+RRR} extends this framework by embedding the regression objective within a Spiking Neural Network training loop, enforcing biological plausibility (e.g., separate E/I populations) on the learned subspace.

\subsubsection{Multi-View Alignment}

Multi-view methods mitigate dependency on session-specific dynamics, which mainly follow the dictionary learning paradigm. To capture the underlying dynamical process, \textit{CREIMBO} \cite{CREIMBO} employed multiple ensemble compositions to decompose the observed activity $\mathbf{Y}^{d}$ into session-specific factors. Based on it, the model formulated the latent trajectory $\mathbf{X}^{d}$ using a shared library of dynamics:
\begin{align}
  \mathbf{Y}^{d} &\approx \mathbf{A}^{d}\mathbf{X}^{d}, \quad \mathbf{X}^{d}_{t+1} = \mathbf{F}^{d}_{t}\mathbf{X}^{d}_{t}, \\
  \mathbf{F}^{d}_{t} &= \sum_{k=1}^{K} c^{d}_{k,t}\mathbf{f}_{k}.
\end{align}
The extracted features $\{\mathbf{f}_{k}\}_{k=1}^{K}$ could then serve as a dictionary of latent sub-circuits invariant across sessions. However, the acquisition of shared motifs is not always straightforward, and the framework effectively captures how different brain regions recruit these universal dynamical motifs over time via the coefficients $c^{d}_{k,t}$.

\section{Behavior-Aligned Latent Modeling}



\subsection{Disentangled Dynamic Modeling}

The core hypothesis here is that the full latent state $\mathbf{x}_t$ can be factorized into orthogonal subspaces: $\mathbf{x}_t = [\mathbf{x}^{beh}_t; \mathbf{x}^{null}_t]$, where only $\mathbf{x}^{beh}_t$ drives the target behavior.

\subsubsection*{Linear Subspace Disentanglement}
\textit{PSID} (Preferential Subspace Identification) \cite{PSID} pioneered this approach by redefining the state-space objective. Instead of maximizing variance explained in neural activity $\mathbf{y}$, it identifies a subspace that maximizes the mutual information between past neural activity and future behavior $\mathbf{z}$. This is achieved via a two-stage projection:
\begin{equation}
\hat{\mathbf{z}}_f = \mathbf{z}_f \mathbf{y}_p^T (\mathbf{y}_p \mathbf{y}_p^T)^{-1} \mathbf{y}_p,
\end{equation}
where $\mathbf{y}_p$ is past activity and $\mathbf{z}_f$ is future behavior. \textit{IPSID} \cite{IPSID} extends this to distinguishing internal dynamics from external sensorimotor inputs $\mathbf{u}_t$, preventing the model from misinterpreting stimulus-driven activity as intrinsic computation.

\subsubsection*{Non-linear \& RNN-based Disentanglement}
To capture complex non-linear dynamics, \textit{DPAD} \cite{DPAD} employs a dual-objective RNN architecture. It learns a non-linear mapping by explicitly partitioning the latent space into behavior-aligned and behaviorally irrelevant dimensions, penalizing the latter to ensure they do not encode action-related information. \textit{BRAID} \cite{BRAID} advances this framework by integrating a forward predictive objective to resolve causal structure. Its generative model explicitly accounts for external control signals:
\begin{equation}
    \mathbf{x}_{t+1}^{beh} = \mathbf{A}_{\psi}(\mathbf{x}_t^{beh}) + \mathbf{K}_{\phi}(\mathbf{u}_t) + \mathbf{w}_t,
\end{equation}
thereby forcing the model to distinguish how inputs $\mathbf{u}_t$ causally drive the behavioral state, separate from internal residual dynamics. Furthermore, \textit{LINT} \cite{LINT} enhances interpretability by imposing low-rank constraints on the RNN connectivity matrix. In contrast, \textit{SRNN} \cite{SRNN} focuses on temporal disentanglement, utilizing switching dynamics to identify discrete cognitive states (e.g., ``planning'' vs. ``execution''), effectively isolating distinct modes of neural computation.

\subsection{Representation Alignment and Geometry}
\subsubsection*{Contrastive Learning \& Geometric Structure}
\textit{CEBRA} \cite{CEBRA} represents a paradigm shift from generative to discriminative modeling. It employs contrastive learning to shape the latent space, pulling together neural time-points that share the same behavioral label (positive pairs) while pushing apart others. The InfoNCE-based objective is:
\begin{equation}
    \mathcal{L} = \mathbb{E}\left[-\psi(\mathbf{x}_i, \mathbf{x}_i^+) + \log \sum_{j=1}^{K} e^{\psi(\mathbf{x}_i, \mathbf{x}_j)}\right].
\end{equation}
This allows CEBRA to align manifolds across different animals performing the same task, even if the underlying neurons are completely different. Complementing this, \textit{GLUE} \cite{GLUE} introduces a geometric metric (i.e., Manifold Capacity) to quantify the efficiency of this encoding:
\begin{equation}
    \alpha_{\mathrm{glue}} = \frac{\Psi_{\mathrm{glue}} \cdot (1 + R_{\mathrm{glue}}^{-2})}{D_{\mathrm{glue}}},
\end{equation}
providing a standardized benchmark for comparing how distinct neural populations encode identical information.

\subsubsection*{Cross-Scenario Consistency Alignment}
Experimental conditions often drift over time. \textit{NLA} \cite{NLA} tackles temporal misalignment (jitter) using a differentiable time-warping function integrated into a contrastive objective. \textit{SABLE} \cite{SABLE} addresses distribution shift using unsupervised Domain Adaptation. It employs a gradient reversal layer to penalize the encoder if it retains session-specific information (e.g., recording artifacts), effectively learning a domain-invariant latent space robust to non-stationarity.

\subsection{Generative and Foundation Models}
Generative and Foundation Models overcome subject-specific training by learning universal representations: large-scale pre-training across diverse datasets addresses data scarcity and zero-shot decoding, interpreting neural activity from unseen subjects or tasks without retraining. We distinguish two categories:

\subsubsection{Deep Generative Models}

The first category focuses on deep generative priors, with a particular emphasis on \textit{Diffusion-Based Models}. \textit{LDNS} \cite{LDNS} combines structured state-space models with Latent Diffusion; it first compresses spikes into a continuous latent space, then trains a diffusion model to generate these trajectories. \textit{BeNeDiff} \cite{BeNeDiff} extends this to video diffusion, serving as an interpretability tool. It visualizes the behavioral meaning of a latent neuron by generating video frames conditioned on specific latent values using classifier-free guidance:
\begin{equation}
    \hat{\boldsymbol{\epsilon}}_\theta(\mathbf{z}_t, \mathbf{c}) = (1+w)\boldsymbol{\epsilon}_\theta(\mathbf{z}_t, \mathbf{c}) - w\boldsymbol{\epsilon}_\theta(\mathbf{z}_t, \emptyset),
\end{equation}
where $\mathbf{c}$ represents the behavioral condition and $w$ scales the guidance strength.

\textcolor{black}{GNOCCHI \cite{GNOCCHI} extends the InfoDiffusion framework to neural time-series data, aiming to infer disentangled latent behavioral variables from recorded neural activity in an unsupervised manner. The method utilizes an auxiliary variable encoder to extract raw neural activity into a latent code $\mathfrak{c}$ regularized via Maximum Mean Discrepancy (MMD); subsequently, a noise prediction network performs conditional diffusion generation conditioned on this code. To guarantee high-fidelity generation while preserving the underlying behavioral structure, the model jointly optimizes the score-matching loss $\mathcal{L}_{\text{score}}$ and the direct data reconstruction loss $\mathcal{L}_{\text{recon}}$:
\begin{equation}
\begin{aligned}
\mathcal{L}_{\text {score }} & =\mathbb{E}\left[\left\|\epsilon_i-\tilde{\epsilon}_i\right\|_2^2\right] \\
\mathcal{L}_{\text {recon }} & =\mathbb{E}\left[\left\|\mathbf{x}_0-\hat{\mathbf{x}}_0\right\|_2^2\right]
\end{aligned}
\end{equation}
where $\tilde{\epsilon}_{i}$ is the predicted noise and $\hat{\mathbf{x}}_0$ represents the reconstructed neural signal sample derived from the predicted noise. Once trained, researchers can conditionally generate high-quality neural activity for unseen behavioral conditions simply by performing a linear traversal within the disentangled latent code space.}

Beyond diffusion, other generative frameworks enforce structural constraints. \textit{Masked VAE} \cite{MaskedVAE} unifies dimensionality reduction with imputation by training on randomly masked spike bins, forcing the model to learn robust joint statistics. Similarly, the \textit{Multiscale Dynamical Model} \cite{MultiscaleDynamicalModel} integrates disparate data types—discrete spikes (millisecond scale) and continuous Local Field Potentials (LFP, second scale)—within a hierarchical EM framework to identify shared latent states that bridge these temporal resolutions.

\subsubsection{\textcolor{black}{Energy-Based \& Autoregressive Models}}

\textcolor{black}{To eliminate the prohibitive iterative inference latency of diffusion models while maintaining high-fidelity neural spiking statistics, \textit{EAG} \cite{EAG} introduces an Energy-based Autoregressive Generation framework to model neural population dynamics in continuous latent spaces without explicit likelihood computation. To preserve the stochastic nature of trial-to-trial neural variability, EAG optimizes an energy-based Transformer via the Energy Score, a strictly proper scoring rule defined as:
\begin{equation}
S(p_{\theta},\textbf{z}_{data})=\mathbb{E}[||\textbf{z}_{1}-\textbf{z}_{2}||^{\alpha}]-2\mathbb{E}[||\textbf{z}-\textbf{z}_{data}||^{\alpha}],
\end{equation}
where $\textbf{z}_1, \textbf{z}_2, \textbf{z} \sim p_\theta$ are independent samples generated by the model, and $\alpha \in (0,2)$ ensures strict propriety to accurately capture single-trial neural statistics.}

\subsubsection{Neuro-Foundation Models}

The third category comprises Neuro-Foundation Models, which prioritize cross-subject generalization. \textit{POYO} \cite{POYO} treats neural recordings as a sequence modeling task, utilizing a unit embedding tokenization scheme:
\begin{equation}
    \mathbf{h}_i = \text{UnitEmbed}(\text{neuron ID}) + \text{RoPE}(t_i),
\end{equation}
which allows the model to process arbitrary sets of neurons from different animals. By leveraging Cross-Attention Transformer blocks, POYO learns a universal mapping between neural tokens and behavioral variables.
\textcolor{black}{To alleviate the high computational complexity and latency constraints of such pure Transformer architectures in real-time, closed-loop settings, \textit{POSSM} \cite{POSSM} introduces a hybrid sequence framework. POSSM retains POYO's single-spike tokenization and cross-attention mechanism to flexibly align neuron identities across diverse sessions, but replaces the heavy Transformer backbone with a modern recurrent state-space model. This modification achieves up to a 9$\times$ inference speedup while preserving powerful cross-subject generalization, scaling models for clinical deployment.}

\textit{NEDS} \cite{NEDS} unifies encoding (behavior $\to$ brain) and decoding (brain $\to$ behavior) into a single Transformer. It employs a multi-task masking strategy (i.e.,randomly masking neural tokens, behavioral tokens, or cross-modal segments) to force the model to learn a joint, generalized representation. This pre-training paradigm allows NEDS to fine-tune on a new animal with limited data, effectively overcoming the data scarcity bottleneck inherent in individual experiments.

\section{Datasets and Benchmarks}


\subsection{Standardized Benchmarks}
To overcome the inconsistencies of ad-hoc evaluations, the field has adopted common benchmarking platforms. A foundational effort is the \textit{Neural Latents Benchmark} (NLB) \cite{pei2021neural}, which evaluates models on standardized, held-out test sets rather than custom splits, providing rigorous and comparable metrics.
Subsequent initiatives have expanded this framework to specialized task domains. While the commonly used monkey hand reaching task serves as a standard baseline, it is often ill-suited for isolating specific sensory feedback mechanisms. The {Area2-Bump} benchmark addresses this limitation by challenging models to explicitly decouple somatosensory perturbations. Similarly, addressing internal cognitive processes where timing evolves dynamically over a single trial, {DMFC-RSG} evaluates the capacity to capture non-motor dynamics and temporal interdependencies. Finally, {Jenkins-Maze} focuses on algorithmic efficiency, testing the scalability of strategies to meet the requirements of large-scale neural recordings.

\subsection{Large-Scale Public Datasets}

\subsubsection{Foundation-Scale Corpora}
The first category, Foundation-Scale Corpora, focuses on massive, multi-region coverage. The \textit{International Brain Laboratory (IBL)} \cite{international2025brain} provides a massive corpus containing Neuropixels recordings from 279 brain regions across 139 mice. With over 600,000 neurons, it serves as the primary substrate for training Neuro-Foundation Models like POYO \cite{POYO} and NEDS \cite{NEDS}. Similarly, the \textit{Allen Brain Observatory} offers extensive datasets combining large-scale Neuropixels and calcium imaging recordings from multiple visual areas in head-fixed behaving mice. While their \textit{Neuropixels} dataset is the standard for benchmarking multi-region communication \cite{siegle2021survey}, the \textit{2-Photon} dataset facilitates geometric alignment testing due to its massive population count of over 63,000 neurons \cite{de2020large}.

\subsubsection{Task-Specific Repositories}
The second category is Task-Specific Repositories. These platforms, such as the \textit{CRCNS} repository, play a crucial role by aggregating datasets contributed by individual laboratories using different recording modalities, animal models, and behavioral tasks. 
Unlike foundation corpora, these datasets often focus on specific dynamical questions. For instance, the {M1/PMd Reach} (pmd-1) dataset captures neural activity from the motor and premotor cortices during multi-target reaching tasks, providing a robust testbed for continuous dynamical systems modeling (e.g., LFADS). Taking a different perspective, the {Hippocampus} (hc-11) dataset records activity during spatial maze navigation; its distinct place-cell activity patterns make it ideal for testing discrete state-switching models (e.g., rSLDS) that track rapid transitions in cognitive representation.
These datasets also have limitations: the prevalence of oversimplified tasks in head-fixed animals likely fails to capture the complexity of neural dynamics during naturalistic, freely moving behavior.

\section{Discussion and Open Challenges}
\label{sec:discussion}

Below, we synthesize fundamental gaps and promising avenues for future development.

\paragraph{Identifiability and Interpretability.}
A central challenge is \textit{identifiability}: guaranteeing that learned parameters match true biological mechanisms rather than merely fitting the data. Deep models often suffer from rotational ambiguity, where latent axes fail to map onto distinct cognitive processes. Constrained approaches such as {Identifiable VAEs} (iVAEs) \cite{khemakhem2020variational} use auxiliary variables to make learned factors biologically unique.

\paragraph{From Correlation to Causal Circuits.}
Current methods are fundamentally observational, struggling to separate causal influence from unobserved ``common drive'' or to predict responses to perturbations. The field must advance toward Causal Representation Learning via in-silico {``Digital Twins''} \cite{LDNS} that test counterfactual hypotheses---e.g., the outcome of optogenetic interventions---before physical experimentation.

\paragraph{Generalization and the ``Universal Neural Grammar''.}
Neuro-Foundation Models face idiosyncratic neural codes that lack a standardized cross-subject vocabulary, preventing zero-shot transfer. The solution lies in {In-Context Learning} and geometric alignment \cite{CEBRA}: treating neural dynamics as a relative manifold rather than absolute firing rates, so models calibrate to new subjects dynamically.

\paragraph{Computational Efficiency.}
Real-time brain-computer interfaces are hindered by the cost of Transformer and probabilistic models. {Knowledge Distillation} into lightweight ``Student'' networks, e.g. Spiking Neural Networks, enables efficient inference on low-power hardware.

\section{Conclusion}

In this paper, we have presented a comprehensive taxonomy of LVMs in neuroscience based on the interdependencies between three pillars: Single-Region Latent Dynamics, Multi-Region Communication, and Behavior-Aligned Modeling. The unified view we adopt in this survey relates hitherto unconnected strands of research, tracing the evolution from static smoothing to nonlinear dynamical systems and distributed information routing. Finally, we also outline open challenges regarding causality, interpretability, and geometry that must be addressed to successfully reverse-engineer the algorithms of biological intelligence.

\section*{Acknowledgements}
This work was partially supported by the National Natural Science Foundation of China (Grant No. 62506090); the National Key R\&D Program of China (Grant No. 2025YFF0523900); an AI2050 Senior Fellowship, a Schmidt Sciences program; the National Science Foundation (NSF); the National Institute of Food and Agriculture (USDA/NIFA; 2023-67021-39829); and the Air Force Office of Scientific Research (AFOSR; FA9550-23-1-0322).

\makeatletter
\let\oldthebibliography\thebibliography
\renewcommand{\thebibliography}[1]{%
  \oldthebibliography{#1}%
  \setlength{\itemsep}{0pt plus 0.3pt}%
  \setlength{\parsep}{0pt}%
  \fontsize{9}{9.3}\selectfont}
\makeatother
\bibliographystyle{named}
\bibliography{ijcai26}

\clearpage
\appendix

\section{Comparison of Representative Methods}
\label{app:comparison}
Table~\ref{tab:comparison} maps the surveyed methods onto the components of the unified generative process of Eqs.~(\ref{eq:dynamics})--(\ref{eq:observation}): the latent dynamics function $f$, the observation model $g$, and the training objective. This structured view complements Figure~\ref{fig:taxonomy} and exposes the design axes along which methods differ. We deliberately compare \emph{modeling assumptions} rather than reproducing single-number leaderboard scores, because reported metrics differ across datasets, train/test splits, and preprocessing pipelines; readers should consult the original papers and the Neural Latents Benchmark~\cite{pei2021neural} leaderboard for head-to-head quantitative results under matched conditions.

\begin{table*}[ht]
\centering
\scriptsize
\setlength{\tabcolsep}{4pt}
\renewcommand{\arraystretch}{1.15}
\begin{tabular}{@{}l l l l l@{}}
\toprule
\textbf{Method} & \textbf{Pillar} & \textbf{Latent dynamics $f$} & \textbf{Observation $g$} & \textbf{Training objective} \\
\midrule
GPFA \cite{GPFA}        & Single   & GP smoothness prior        & Gaussian linear      & Marginal likelihood (EM) \\
vLGP \cite{vLGP}        & Single   & GP prior                   & Poisson              & Variational inference \\
PLDS \cite{PLDS}        & Single   & Linear ($\mathbf{A}$)      & Poisson              & Likelihood (Laplace--EM) \\
LFADS \cite{LFADS}      & Single   & Generator RNN              & Poisson              & Sequential VAE (ELBO) \\
NDT \cite{NDT}          & Single   & Transformer attention      & Poisson              & Masked modeling \\
PLNDE \cite{PLNDE}      & Single   & Neural ODE                 & Poisson              & ELBO \\
LangevinFlow \cite{LangevinFlow} & Single & Langevin SDE (oscillators) & Poisson      & Sequential VAE (ELBO) \\
rSLDS \cite{rSLDS}      & Single   & Switching linear           & Gaussian/Poisson     & Bayesian inference \\
gpSLDS \cite{gpSLDS}    & Single   & GP-switching linear        & Poisson              & Variational inference \\
DLAG / mDLAG \cite{DLAG,mDLAG} & Multi & Delayed GP                 & Gaussian             & Variational EM \\
ADM \cite{ADM}          & Multi    & Time-varying Markov GP     & Gaussian             & Scalable variational \\
MR-SDS \cite{MR-SDS}    & Multi    & Switching nonlinear        & Poisson              & ELBO \\
MR-LFADS \cite{MR-LFADS}& Multi    & Sender--receiver RNN       & Poisson              & ELBO \\
CCA / RRR \cite{CCA,RRR-based} & Multi & Static linear subspace    & Gaussian (LS)        & Corr.\ / low-rank regression \\
CREIMBO \cite{CREIMBO}  & Multi    & Shared linear dictionary   & Gaussian             & Dictionary learning \\
PSID \cite{PSID}        & Behavior & Linear                     & Gaussian             & Subspace ident.\ (predictive) \\
DPAD \cite{DPAD}        & Behavior & RNN (dual)                 & Poisson/Gaussian     & Prioritized prediction \\
CEBRA \cite{CEBRA}      & Behavior & Encoder (non-generative)   & ---                  & Contrastive (InfoNCE) \\
LDNS \cite{LDNS}        & Behavior & SSM + latent diffusion     & Poisson              & Score matching \\
GNOCCHI \cite{GNOCCHI}  & Behavior & Conditional diffusion      & ---                  & Score + reconstruction + MMD \\
EAG \cite{EAG}          & Behavior & Autoregressive (energy)    & ---                  & Energy score \\
POYO \cite{POYO}        & Foundation & Cross-attention Transformer & ---                & Supervised decoding \\
POSSM \cite{POSSM}      & Foundation & Hybrid SSM + cross-attn   & ---                  & Supervised decoding \\
NEDS \cite{NEDS}        & Foundation & Transformer (multi-task)  & Poisson/Gaussian     & Multi-task masking \\
\bottomrule
\end{tabular}
\caption{Representative methods compared along the axes of the unified generative process (Eqs.~\ref{eq:dynamics}--\ref{eq:observation}). ``---'' denotes discriminative or decoding-oriented models without an explicit observation likelihood.}
\label{tab:comparison}
\end{table*}

\section{Extended Discussion of Open Challenges}
\label{app:discussion}
We expand the four challenges of Section~\ref{sec:discussion} into the gap-versus-avenue form below.

\paragraph{Identifiability and Interpretability.}
\textit{The Current Gap:} Most deep generative models suffer from rotational ambiguity---an affine transformation of the latent space can yield identical data likelihoods while representing completely different biological hypotheses. Consequently, while current methods excel at \textit{fitting} data, the extracted axes rarely map one-to-one onto distinct biological mechanisms (e.g., distinguishing ``planning'' from ``execution'' subspaces).
\textit{Future Avenues:} Future work must pivot from purely unsupervised learning to constrained modeling with strong inductive biases. A promising direction is \textbf{Identifiable VAEs} (iVAEs)~\cite{khemakhem2020variational}, which leverage auxiliary variables (behavioral context or stimulus onset) to statistically constrain the latent space so that learned factors are not just accurate, but biologically unique.

\paragraph{From Correlation to Causal Circuits.}
\textit{The Current Gap:} Subspace identification methods map communication channels but remain observational; they cannot distinguish direct causal influence from ``common drive,'' where two regions appear correlated only because they share input from a third, unobserved source. Standard LVMs describe how a system evolves \textit{naturally} but rarely predict its response to \textit{perturbations} (stimulation or lesions).
\textit{Future Avenues:} The field must move toward Causal Representation Learning, validating models on their ability to predict the outcome of interventions. Generative diffusion models~\cite{LDNS} offer a path toward in-silico \textbf{Digital Twins} that test counterfactual hypotheses---e.g., the effect of an optogenetic perturbation---before physical experimentation.

\paragraph{Generalization and the ``Universal Neural Grammar''.}
\textit{The Current Gap:} Unlike words in natural language, neural codes are idiosyncratic; the ``meaning'' of a spike depends on the animal, the region, and even daily electrode drift, creating a hard barrier to zero-shot transfer across subjects.
\textit{Future Avenues:} Solutions lie in In-Context Learning and geometric alignment. Architectures that treat neural activity as a relative manifold rather than absolute firing rates~\cite{CEBRA}, combined with calibration ``prompts'' at inference, will be essential for universal decoders.

\paragraph{Computational Efficiency.}
\textit{The Current Gap:} Probabilistic models often scale cubically with sequence length and Transformers quadratically, which is prohibitive for closed-loop brain--computer interfaces that demand millisecond-level feedback.
\textit{Future Avenues:} Knowledge distillation of large ``Teacher'' models into lightweight ``Student'' networks---State Space Models or Spiking Neural Networks---can preserve inference quality on low-power, implantable hardware. The hybrid state-space design of POSSM~\cite{POSSM} is an early step in this direction.

\section{Mapping Subspace and Discriminative Methods to the Unified Framework}
\label{app:framework}
The generative process of Eqs.~(\ref{eq:dynamics})--(\ref{eq:observation}) is most natural for state-space and deep generative models, but the subspace and contrastive methods we survey can also be read through it. \textbf{Communication-subspace methods} (CCA~\cite{CCA}, RRR~\cite{RRR-based}) correspond to the static, single-step special case in which $f$ is dropped and $g$ is a low-rank linear map between two simultaneously observed populations; the ``objective'' is correlation maximization or rank-constrained least squares rather than likelihood, so these methods estimate the \emph{observation} geometry without an explicit dynamics prior. \textbf{Contrastive methods} such as CEBRA~\cite{CEBRA} are discriminative: they do not posit an observation density $\mathcal{P}(g(\cdot))$ and instead learn an encoder $q(\mathbf{x}_t\mid\mathbf{y}_t)$ trained by InfoNCE so that the latent geometry aligns with auxiliary labels. Placing them in the same diagram clarifies that the framework's three knobs---$f$, $g$, and the objective---degenerate gracefully: setting $f$ to identity recovers static subspace models, and replacing the generative objective with a contrastive one recovers representation-alignment methods.

\section{Relation to Prior Surveys}
\label{app:related}
Earlier reviews are complementary to ours but differ in scope. Panahi et al.~\cite{Panahi2022} survey generative models of brain dynamics across biophysical, phenomenological, and data-driven categories, emphasizing mechanistic simulation rather than latent-variable inference from population recordings. Zhou et al.~\cite{Zhou2025BFM} review brain foundation models with a focus on large-scale pretraining and multimodal neural signal processing. Our survey is narrower and more methodological: it unifies latent-variable models for population electrophysiology specifically across single-region dynamics, multi-region communication, and behavior alignment, organizing them under one generative process. We therefore frame our contribution as the first treatment to connect these three threads within a single comparative framework, rather than as the first survey of neural generative models in general.

\section{Additional Method Classes and Recording Modalities}
\label{app:classes}
For completeness we note directions that fall outside our three pillars but are active alternatives.
\textbf{Normalizing flows} provide exact likelihoods and invertible maps for characterizing manifold geometry and curvature~\cite{NormFlowManifold}, complementing the variational and diffusion priors emphasized here.
\textbf{Graph-based connectivity} methods learn explicit neuron-to-neuron or region-to-region directed functional graphs, an alternative to the latent-subspace view of multi-region communication.
\textbf{Topological data analysis} compares the global shape of neural manifolds across populations and tasks.
On the \textbf{encoding} direction (behavior/stimulus $\rightarrow$ neural activity), self-supervised video models such as BEAST~\cite{BEAST} can rival contrastive encoders; our Behavior-Aligned pillar covers decoding and joint encoding--decoding (NEDS~\cite{NEDS}) but not pure encoding.
Finally, several covered methods are validated mainly on spike trains: Gaussian-process delay models such as DLAG/mDLAG can be numerically unstable on non-Gaussian intracranial LFP, motivating modality-robust variants~\cite{SPIRE}.


\end{document}